# Development and Transcription of Assamese Speech Corpus

Himangshu Sarma, Navanath Saharia, Utpal Sharma, Smriti Kumar Sinha, Mancha Jyoti Malakar
Department of Computer Science and Engineering
Tezpur University
Napaam, India – 784028
himangshu.tezu@gmail.com
{nava_tu,utpal,smriti}@tezu.ernet.in
manchajyotiassam@gmail.com

*Abstract*—A balanced speech corpus is the basic need for any speech processing task. In this report we describe our effort on development of Assamese speech corpus. We mainly focused on some issues and challenges faced during development of the corpus. Being a less computationally aware language, this is the first effort to develop speech corpus for Assamese. As corpus development is an ongoing process, in this paper we report only the initial task.

## I. INTRODUCTION

Language analysis is a very broad area of research till now. Every speech can be represented phonetically by some finite set of symbols which are called phonemes of the language. In this paper we report how we collect a speech corpus and how we transcribe the speech corpus. We report the different issues faced during these two steps. Initially we identify the phonemes used during our transcription.[1] Almost four hours of transcription done using International Phonetic Alphabet 2005 (IPA).[2] We use some tools for our transcription. During the transcription we find that from the IPA 2005 some of phonemes are absent in Assamese language. From our collected corpus we find that some of Assamese words are more frequently used by Assamese speakers than other words. During transcription we find some interesting observation and some issues about the Assamese language. The Assamese speech corpus is collected from different people of different regions of Assam. So, from our collected speech corpus we find that sometimes the same word is pronounced differentially by speakers of different regions. We find that the pronunciation of many speakers of Assamese are influenced by other languages, such as English, Hindi and Bengali. Phones not originally present in Assamese often occur in Assamese speech.

We structure the paper as: In Section II we describe how we collected the speech data. In Section III we discuss the features of Assamese language and phonemes. In Section IV we analyzed about different tools used in our transcription. In Section V we report the frequency analysis of different phonemes of Assamese. In Section VI we describe about different issues and some interesting observation during transcription. In Section VII we state our conclusion and possible further work in this line. Acknowledgment to DIT has been given in the Section VIII of this proposed paper.

## II. DATA COLLECTION

We have collect almost seven hours Assamese speech with more than twenty five speakers of different regions of Assam. The following are the information of our Assamese speech corpus.

### A. Mode of Recording

To collect the Assamese speech corpus we collect the recorded Assamese speech in different categories. We collect Assamese speech in three categories.

a) Conversation Speech
b) Extempore Speech
c) Read Speech

Under read speech we collect speech in two categories.
i) Air News Speech
ii) Text Read Speech

### B. Environment of Recording

To build a speech corpus the recording environment is very important. To collect our Assamese Speech Corpus when we recorded our speech from different speaker in a closed room.

### C. Device Used for Recording

To collect our Assamese speech corpus we record our data using **SONY ICD-UX533F**, which is a hand held recording device.

## D. Encoding
To record our data we follow some encoding rule. Speech data are recorded at 41 KHz sampling frequency.

## E. Age Group
Recorded data of the Speech data is collected from people of 20-40 years.

## F. Mother Tongue
All data of the Assamese speech corpus is collected by Assamese native speakers.

## G. Male Female Ratio
In our Speech Corpus we have recorded speech from 14 male and 11 female.

## III. FEATURES OF ASSAMESE LANGUAGE AND PHONEMES

There are 41 languages in North-East India only.[3] Where Assamese is the one of the most spoken language of North-East. Assamese is an Eastern Indo-Aryan language.[4]

In Assamese language Labiodental, Dental, Retroflex, Uvular, Pharyngeal, Trill, Lateral fricative phoneme are not present.[5]

## A. Consonants
In the transcription of Assamese language we use twenty six phonemes from International Phonetic Alphabet 2005 to define Assamese consonants.

## B. Vowels
We used nine phonemes from International Phonetic Alphabet 2005 for our Assamese vowels to transcribe our Assamese Speech.

## C. List of Consonants and Vowels
In the following table we define Assamese consonants and vowels with corresponding IPA phoneme used in our transcribed speech of our speech corpus.

**TABLE I. Consonant and Vowel with corresponding IPA Phoneme**

| | LETTER | IPA | LETTER | IPA | LETTER | IPA |
|---|---|---|---|---|---|---|
| | ক | /k/ | ণ | /n/ | ৱ | /ɯ/ /bɒ/ |
| | খ | /kʰ/ | ত | /t/ | শ | /x/ |
| C O N S O N A N T S | গ | /g/ | থ | /tʰ/ | ষ | /x/ |
| | ঘ | /gʰ/ | দ | /d/ | স | /s/ /x/ |
| | ঙ | /ŋ/ | ধ | /dʰ/ | হ | /h/ |
| | চ | /ʃ/ /ʧ/ | ন | /n/ | ক্ষ | /kʰj/ |
| | ছ | /ʃ/ /ʧ/ | প | /p/ | য | /j/ |
| | জ | /dʒ/ | ফ | /pʰ/ | ড় | /ɹ/ |
| | ঝ | /dʒʰ/ | ব | /b/ | ঢ় | /ɹh/ |
| | ঞ | /ɲ/ | ভ | /bʰ/ | ৎ | /t̪/ |
| | ট | /t/ | ম | /m/ | ○ং | /ŋ/ |
| | ঠ | /tʰ/ | য | /dʒ/ | ○ঃ | ▼ |
| | ড | /d/ | ৰ | /ɹ/ | ◌̐ | ~ |
| | ঢ | /dʰ/ | ল | /l/ | | |
| V O W E L S | অ | /ɒ/ /ɔ/ | উ | /u/ | ঐ | /oi/ |
| | আ | /a/ | ঊ | /u/ | ও | /ʊ/ /o/ |
| | ই | /i/ | ঋ | /ɹi/ | ঔ | /oʊ/ |
| | ঈ | /i/ | এ | /e/ /ɛ/ | | |

## IV. TOOLS USED DURING TRANSCRIPTION
To transcribe our Assamese Speech Corpus using IPA symbols, we mainly used two tools. They are :

### A. Wavesurfer
It is used to check the waveform, spectrogram to find out correct transcription of the speech.[6]

### B. IPA Typing Tool
To transcribe the speech we need the IPA fonts. IPA fonts are not available in general keyboard. So, we used IPA typing tool to transcribe our speech.[7]

## V. FREQUENCY ANALYSIS
We find the frequencies of all thirty five IPA phonemes from our more than four hour IPA transcription of Assamese recorded data of twenty five speakers from different regions of Assam.

TABLE II. Phoneme frequency list

| Phonems | Frequency | Phonems | Frequency |
|---|---|---|---|
| p | 2073 | pʰ | 123 |
| b | 3684 | bʰ | 567 |
| t | 5230 | tʰ | 810 |
| d | 1691 | dʰ | 481 |
| k | 3745 | kʰ | 871 |
| g | 657 | gʰ | 107 |
| m | 2691 | n | 4096 |
| ɲ | 73 | ŋ | 350 |
| ɹ | 7199 | s | 531 |
| ʃ | 1975 | x | 2643 |
| h | 1755 | j | 1535 |
| ɰ | 1453 | l | 3028 |
| tʃ | 6 | ʤ | 1741 |
| a | 9567 | e | 1204 |
| i | 9608 | o | 272 |
| u | 3905 | ɛ | 3706 |
| ɔ | 1892 | ɒ | 11302 |
| ʊ | 1731 | | |

### A. Most And Less Frequent Word

During our transcription we find that some of words are most frequently used and some are very less frequent. In the following table we show five most and least frequent words found during our transcription.

TABLE III. Most and Less Frequent Word

| MOST FREQUENT | | LESS FREQUENT | |
|---|---|---|---|
| WORDS | IPA | WORDS | IPA |
| আৰু | aɹu | সদভাৱ | xɒdɒbʰabɒ |
| এই | ei | অংশটোত | ɒŋxɒtot |
| কৰি | kɒɹi | মণ্ডৰ | mɒndɒɹ |
| পৰা | pɒɹa | সাধনত | xadʰɒnɒt |
| হৈ | hoi | বিশ্বজনীনতাৰ | bixbɒʤɒninɒtaɹ |

### B. Most Frequent Phoneme

From TABLE III we found that ɹ is the highest used consonant in our transcription which is used for Assamese word . e.g.

রাখে = ɹakʰe
ৰাইজক = ɹaiʤɒk
ৰাজ্য = ɹaʤjɔ
সুৰাপায়ী = xuɹapaji
বাতাবৰণ = batabɒɹɒn
কৰিব = kɒɹibɒ
লোকৰ = lukɒɹ
খবৰ = kʰɒbɒɹ
মোৰ = muɹe

We found that ɒ is the most used vowel phoneme during our transcription. ɒ is used for Assamese letter . e.g.

অনা = ɒna
ভিতৰত = bʰitɒɹɒt
বৃহস্পতিবাৰ = bɹihɒspɒtibaɹɒ

### C. Less Frequent Phoneme

From the transcription we found that ɲ is the less used consonants which is shown in TABLE III. It is used for . e.g.

চিঞৰ = ʃiɲɒɹ
মঞ্চত = mɒɲʃɒt

In the vowel section we found that o is less used vowel in our transcription database. e.g.

সাহিত্যিকো = xahitjiko
দৈনন্দিন = doinɒndin

## VI. ISSUES AND INTERESTING OBSERVATIONS

During transcription we face different problems and some interesting observation related to consonants and vowels. The pronunciation rules are different from different regions of Assam. The following are the some interesting observation and problems faced during the transcription.

## A. Consonants

1) **w** is used for if is used in middle of a word or the first letter of the word is but **bɒ** is used for if is the last letter of the word. e.g.

    ছোৱালী = ʃoɯali
    ঘৰুৱা = gʰɒɹuɯa
    ৱাহিদ = ɯahid
    ৱাল = ɯal
    দেৱ = debɒ
    কেশৱ = kexɒbɒ

2) **ʧ** is actually not used in proper Assamese pronunciation. But **ʧ** is used in English, Bengali and Hindi languages. Now a days maximum Assamese people speak English, Bengali and Hindi. So, when those people speak Assamese they used **ʧ**. e.g.

    আচ্ছা = aʧa
    বাচ্চা = baʧa

3) During transcription we saw that there is a **ɒ** occurs after every consonant. But some words i.e. , , , are not follow this rule. e.g.

    অংক = ɒŋkɒ
    আঙুৰ = ɒŋuɹ
    উৎসৱ = ut̯xɒb
    নিঃকিন = nih̯kin

4) We find another one interesting observation for . When we used single than we used **x**. But when we used at clustering we used **s**. e.g.

    সাধাৰণ = xadʰaɹɒn
    সাবতি = xabɒti
    ব্যৱস্থাত = bjbɒstʰat

5) We also find that in some consonant both sequence happens. In some words **ɒ** occurs after the consonant and in some words **ɒ** does not occurs after the consonant. e.g.

    ৰাইজক = ɹaidʒɒk
    জগতৰ = dʒɒgɒtɒɹ
    ৰাজপথ = ɹadʒpɒtʰ
    ৰাজগড় = ɹadʒgɒɹ

## B. Vowels

**ʊ** and **u** is sometimes placed for same word which meaning is same, but placed different phoneme because of different region of speakers. e.g.

    বোলেও = bʊleo
    বুলেও = buleo
    তোমাৰ = tʊmaɹ
    তুমাৰ = tumaɹ

## C. Clustering

If a consonant cluster occurs in the final position of a word then the vowel **ɒ** insert at the last position of the word. e.g.

    মন্তব্য = mɒntɒbjɒ
    অস্তিত্ব = ɒstitbɒ
    ৰক্ত = ɹɒktɒ

But, in some cases if a consonant cluster occurs final position then **ɒ** is not inserted at the last position of the word. e.g.

    কান্ধ = kandʰ
    বান্ধ = bandʰ

## VII. CONCLUSIONS AND FURTHER WORK

In this paper we present the issues and some interesting observations of development and transcription of Assamese Speech Corpus. We have developed a speech corpus by recording seven hours of speech from twenty five different speakers from different regions of Assam. Out of seven hours we have transcribed almost four hours of data. We have analyzed the phonemes occurrence frequencies and problems faced during transcription. We have made some interesting observations regarding the phonetic transcription and standard written Assamese. In further work of the Assamese speech corpus we are going to collect more different categories of data from more speakers from different. Also in future we are going to work about the syllabification and breakmarking of our speech corpus to have a more useful Assamese speech processing resource.

## VIII. ACKNOWLEDGEMENT

The work reported above has been supported by **DIT** sponsored project Development of Prosodically Guided Phonetic Engine for Searching Speech Databases in Indian Languages letter no [11(6)/2011-HCC(TDIL)].